\documentclass[authoryear,preprint,review,12pt]{elsarticle}

\usepackage{graphicx}
\usepackage[english]{babel}

\usepackage{amssymb,amsmath}
\usepackage{lineno}
\usepackage{natbib}
\usepackage{color}
\usepackage{placeins} % Provide the FloatBarrier command 
\usepackage{multirow}
\usepackage[linesnumbered,boxed,ruled]{algorithm2e}
\usepackage[squaren, Gray, cdot]{SIunits}
\usepackage{datetime}

\settimeformat{ampmtime}
\usepackage{multirow}

\journal{Computer and Electronics in Agriculture}
\DeclareGraphicsExtensions{.png}
\usepackage[T1]{fontenc}

\newcommand{\isa}{{\preceq}}

\newcommand{\hascond}{{\mathcal{HCO}}}

\newcommand{\hasconstr}{{\mathcal{HCS}}}

\newcommand{\conceptset}{{\mathcal{C}}}
\newcommand{\condset}[2]{{\{#1|#2\}}}

\begin{document}

\begin{frontmatter}

\title{A %data and knowledge-based 
modeling approach to design a software sensor and analyze agronomical features - Application  to sap flow and grape quality relationship}

\author[inra]{Aur\'elie Th\'ebault}
\author[fruition]{Thibaut Scholasch}
\author[inra]{Brigitte Charnomordic}
\ead{bch@supagro.inra.fr}
\author[inra]{Nadine Hilgert}
\address[inra]{INRA-SupAgro, UMR 729 MISTEA, F-34060 Montpellier, France}
\address[fruition]{Fruition Sciences, Montpellier, France} 

\begin{abstract}
This work proposes a framework using temporal data and domain knowledge in order to analyze complex agronomical features. The expertise is first formalized in an ontology, under the form of concepts and relationships between them, and then used in conjunction with raw data and mathematical models to design a software sensor. Next the software sensor outputs are put in relation to product quality, assessed by quantitative measurements. This requires the use of advanced data analysis methods, such as functional regression. 
The methodology is applied to a case study involving an experimental design in French vineyards. The  temporal data consist of sap flow measurements, and the goal is to explain fruit quality (sugar concentration and weight), using vine's water courses through the various vine phenological stages. 
The results are discussed, as well as the method genericity and robustness.
\end{abstract}

\end{frontmatter}

\linenumbers

%%%%%%%%%%%%%%%%%%%%%%%%%%%%%%%%%%%%%%%%%%%%%%%%%%%%%%%%%%%%%%%%%%%%%%%%
\section{Introduction \label{sec:intro}}
In modern Agronomy, the recent progress of sensors provides a lot of data, among them many temporal data. This opens new challenges, such as the proper calibration of these sensors, and the use of temporal data to establish relationships with product characteristics and quality. These relationships are not easy to determine because of the high variability of biological material. This can be compensated by the integration of expertise, as Agronomy is a domain that has always relied as much on experience than on science. Nevertheless, for domain knowledge to be effectively used in collaboration with mathematical models and data, an expertise formalization step is required.
%Mathematical models can help in that task, but they are rarely available at the right scale.

 Our objective in this paper is to show the interest of a formalized data and knowledge-based approach to study a complex agronomical phenomenon, namely the influence of vine water deficit on grape quality. 

Any index of vine water deficit aims at evaluating the amount of water effectively used by the vine in order to determine if if this amount falls short of some reference amount. Typically, the \textit{reference} amount is the maximal amount of water a vine can use.

Various methods exist to characterize the level of water deficit experienced by the plant as reviewed by \cite{Jones2004}. Tissue water status can be assessed visually or by measurements of vine water potential. However, both methods have serious drawbacks. The lack of precision of visual observations often leads to yield reduction before visible symptoms occurs.  The pressure chamber method used to measure water potential is slow and labour intensive, especially for predawn measurement, and is unsuitable for automation. In addition, measurements done with pressure chambers are very dependent on atmospheric conditions and vine phenological stage \cite{olivo2009,williams2007,rodrigues2012,Santesteban2009}. 

Sap flow sensors, that indirectly measure changes in conductance, have recently become available. This sensitive measurement method requires a complex instrumentation and technical expertise for the definition of irrigation control thresholds \cite{ginestar1998}. The main advantages of sap flow method is to allow automatic and continuous measurement of water flowing through the plant, which is directly related to transpiration  \cite{escalona2002,Jones2004, Cifre2005, Zhang2011}.

However, sap flow is a complex phenomenon and expert knowledge is necessary to convert raw data into useful transformed data, i.e. water courses, by designing a software sensor. Once these data transformations are validated, it opens the way to a range of new studies, based on vine's water courses. 

In this paper, we will first show how a formalized data and knowledge-based approach can be useful to design a  software sensor.  Knowledge formalization will be done by using ontologies, which  take increasing importance in the field of Life Sciences \cite{Villanueva08, Thomopoulos13}, for their ability to model and structure qualitative domain knowledge.

In a second step, water use trajectories will be put in relation to grape quality indicators such as Berry Weight or Sugar Concentration, using recent data analysis tools and formalized knowledge. Innovative data analysis tools include functional data analysis that offers the possibility to use curve (functional) data instead of scalar data. Functional data analysis has not been much used in life sciences yet \cite{ullah2013}, though it could be of particular interest in the Vine and Wine Industry, and more generally for modern Agronomy.

The modeling task is divided into two independent parts: software sensor design and temporal data analysis. If the sensor design procedure were different, this would not affect the validity of the data analysis methodology.

The methodological work is illustrated by a case study, involving an experimental design on several vineyards in the Languedoc region (France).

The paper is organized as follows: Section \ref{sec:meth} presents the material and methods. It is divided into four parts. The first part gives some elements about data and the second one presents ontology-based formalization. The software sensor design, that relies on the use of mathematical models, data and formalized knowledge,  is  described in the third part.  The selected example shows how it is possible to transform raw sap flow data into vine water deficit courses. The fourth part describes the methods used for analysing  the sofware sensor output in relation to product quality. Section \ref{sec:resu} presents and discusses the results and their relationship with grape composition (Sugar Concentration, Berry Weight).  Some concluding remarks are given in Section \ref{sec:discu}.

% are discussed in Section \ref{sec:discu}, in relation with the relevant literature. In Section  \ref{sec:discu}, a few possible steps towards an increased automation of the approach and its improvements are also discussed.
%%%%%%%%%%%%%%%%%%%%%%%%%%%%%%%%%%%%%%%%%%%%%%%%%%%%%%%%%%%%%%%%%%%%%%%%\\
%%%%%%%%%%%%%%%%%%%%%%%%%%%%%%%%%%%%%%%%%%%%%%%%%%%%%%%%%%%%%%%%%%%%%%%%

\section{Material and methods\label{sec:meth}}
In this section, we propose to follow four steps:
\begin{itemize}
\item to describe the experimental design with its input and output variables
\item to formalize eco-physiological knowledge using an ontology
\item to design a software sensor using formalised knowledge, a mathematical model, and data
\item to relate software sensor output to product quality using decision trees and functional analysis.
\end{itemize}
%Our objective in this paper is to generate a continuous vine water stress indicator and to relate it to grape quality. 
\begin{figure}[hbtp]
\begin{center}
\centering\includegraphics[width=0.85\linewidth]{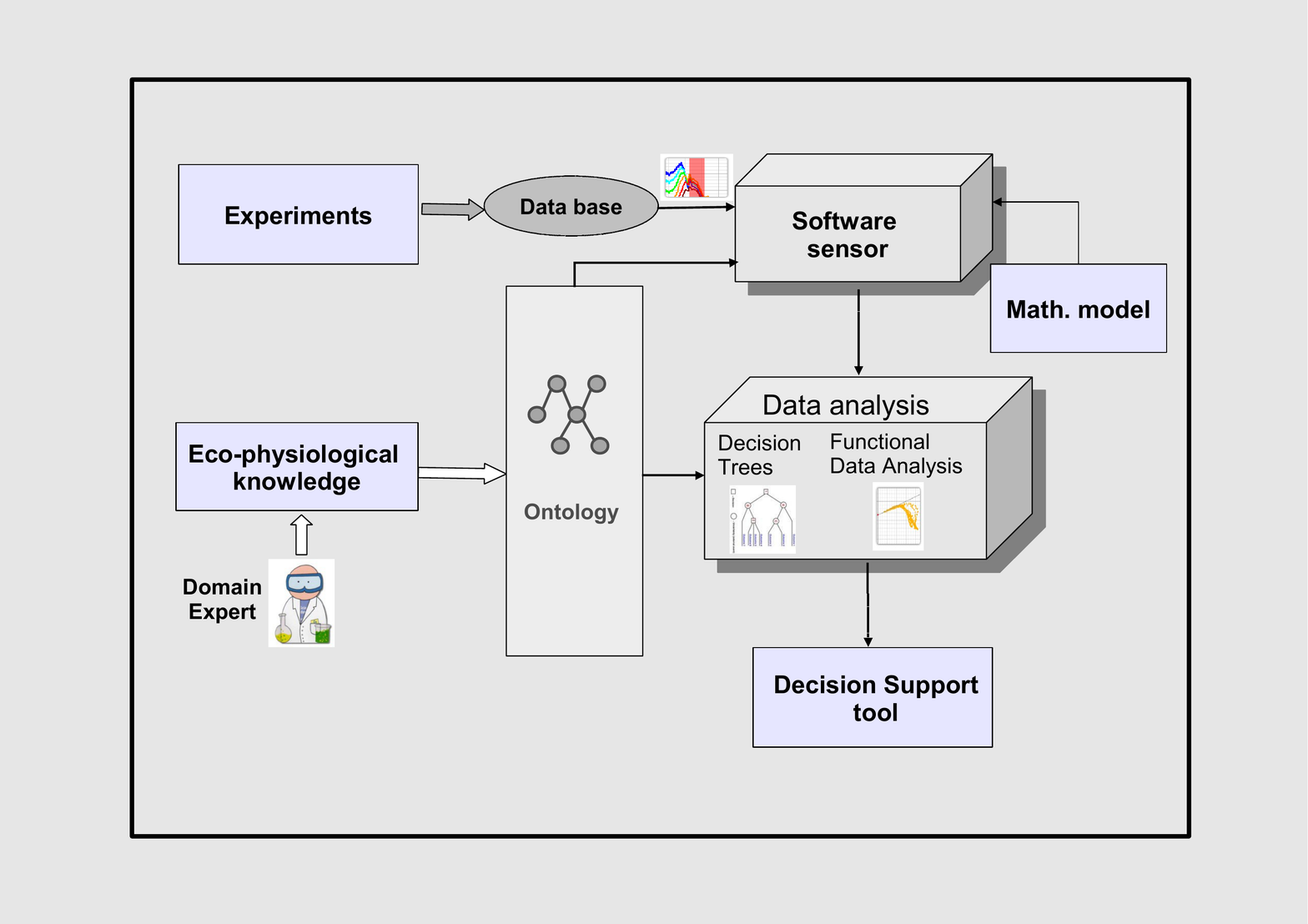}
\caption{Outline of the proposed modeling approach.}
\label{fig:KD}
\end{center}
\end{figure}

Expert knowledge plays an essential part in the modeling process, and we focus on providing an efficient way to separate the data-based statistical procedures from the qualitative knowledge-based assumptions.

The outline of the approach is given in Fig.\ref{fig:KD}. Experiments feed a data base. The software sensor integrates data from the data base, an ontology and a mathematical model. Its outputs can be analyzed using data analysis. This analysis also calls for  eco-physiological knowledge, essentially about the phenological stages. Therefore, the ontology is used at two different levels: software sensor design and data analysis supervision. Data analysis is performed on the basis of two complementary perspectives for determining relations between software sensor temporal output and product quality. The first line of work is to design scalar explanatory variables, by summarizing a period of interest in compliance with the eco-physiological knowledge. These variables can then be used as input to decision trees. The second line of work is to use recent advances in functional data analysis, so that the inputs to the statistical model are the temporal data as a whole.

In the following, the approach is illustrated with a software sensor to estimate vine water courses and their relation to grape quality. Nevertheless, the proposed methodology is generic in many aspects, and could be useful in the development of other decision support tools in Agronomy, provided that expertise and temporal data are available.

%These elements provide the foundation on which our approach is based.

\subsection{Experimental design}\label{sec:design}
Data used in this paper come from a multi-site experiment located in the south of France. The same experimental design was set up in seven sites across the Languedoc Roussillon region in order to test for the effects of vine stress status on grape potential and wine quality in contrasted environmental conditions. In total, vine water stress status was followed over 16 vine plots, each planted with one of the following varieties: Merlot, Cabernet-Sauvignon, Grenache or Chardonnay. 

To get a wider range of vine water statuses during the season, an irrigation treatment was applied for two years on each of the eight site-variety combinations. The irrigation treatment consisted of two modalities, replicated twice, yielding 32 experimental subplots. In the non irrigated subplots, vines only received natural precipitations during the growing season while in the irrigated subplots, vines received regular extra-amounts of water through drippers line (emission rate from 2 to 4 $\litre\ \reciprocal\hour$, 1 to 2 drippers per plant). 

Several kinds of data, collected according to an experimental design are available: local meteorological data, vine water stress related measurements, phenological state assessments, as well as grape quality analyses.

\subsubsection{Meteorological data}\label{sec:meteo}

 %\textit{Raw data}

Hourly meteorological data on windspeed ($\kilo\metre\ \reciprocal\hour$), minimal, maximal and mean air temperature ($\celsius$), air humidity (\%), solar radiation ($\watt\ \meter\rpsquared$) and amounts of precipitations ($\milli\metre$) were extracted from local meteorological stations for each site. 

 \vspace{0.25cm}

\textit{Transformed data}

Hourly vapor pressure deficit ($\mathit{VPD}$) and reference atmospheric evaporative demand (potential evapotranspiration, $\mathit{ET}_{ref}$) were calculated according to methodologies referred to as FAO-56 \cite{Allen1998}.
Calculation of reference atmospheric evaporative demand ($\mathit{ET}_{ref}$ in $\milli\meter\ \reciprocal\dday$) is based on Penman-Monteith formula.

Daily meteorological data were obtained from hourly data after a trapeze integration. Thermal time, {\itshape i.e.} the accumulation of growing degree days (GDD) from April 1$^{st}$ was calculated by daily integration of mean air temperature minus a base temperature of 10 $\celsius$ which is considered as the simplest model to estimate vine phenology \cite{parker2011}.

\subsubsection{Phenological data}\label{sec:pheno}
The main phenological phases (budbreak, bloom, nouaison, veraison) were estimated visually in each experimental plot when 50\% of the plants reached the stage. Bloom was observed when 50\% of the clusters had the cap off. Nouaison was defined using the bloom stage, according to local expert knowledge (see Section \ref{sec:con}).Veraison dates were recorded when 50\% of the fruit had turned red. 
%Based on berry Sugar Concentration and subsequent potential alcoholic degree, a maturity stage was defined for each variety. 

\subsubsection{Vine water status data}\label{sec:Tr}
Vine water status was monitored by two kinds of measurements: discrete measurements of leaf water potential at predawn ($\Psi_{b}$, or predawn LWP) and continuous measurements of sap flow.

\vspace{0.25cm}
\textit{Leaf water potential at predawn}

LWP measurements were conducted every week from the end of June to the end of August with a pressure chamber at predawn (between  $\approx$ 3.00 am  and  $\approx$ 5.00 am). 

\vspace{0.25cm}
\textit{Sap flow}

The energy balance method (Sakuratani, 1981) was used to measure sap flow with Sap IP system (Dynamax, Houston, TX, USA). There is one variety per vineyard site. The vineyard site is divided into 2 irrigation treatments. Two vineyard rows were selected. One row represents one irrigation treatment.  In each selected row, 2 vines were equipped with one sensor. Each sensor measured vine sap flow rate every 15 minutes. The 2 selected vines were within 25 meters of each other within the same row.  

%\vspace{0.25cm}
%\textit{Transformation of sap flow data}

Sap flow rates measured on each vine were averaged on an hourly basis within each row. Total sap flow of each vine was calculated as the product of sap flux density and cross sectional sap wood area at the point of measurement. Various expert methods were applied to filter out nighttime, weak and erroneous signals. Sap flow measurements were scaled at the plant level according to plant leaf area estimates corresponding to each sensor. The daily sap flow assumed to measure daily vine transpiration was computed by adding all hourly sap flow rates measured during the day. The volumetric flux per vine ($\gram.\reciprocal\hour$)  was converted into $\milli\meter.\reciprocal\hour$ taking into account the respective area of ground per vine. 
Daily vine transpiration will be noted $T(t)$.

%Daily transpiration ratios are calculated from daily sap flow rates $Tr$ as follows:
%\begin{equation}\label{eqn:tr}
%Tr_{norm}=\frac{Tr}{ET_{ref}}
%\end{equation}

\subsubsection{Fruit composition quality data}\label{sec:dataAroma}

Starting two weeks before harvest, fruit was sampled for each irrigation treatment in each vineyard. Fruit data was collected at three different dates. Fruit composition analysis focused on berry weight ($\gram$),  
 sugar concentration ($\gram\ \reciprocal\litre$), acidity ($\gram$H2SO4$\ \reciprocal\litre$), anthocyanes and assimilable nitrogen ($\milli\gram\ \reciprocal\litre$).
%Sugar Concentration in grapes and berry weigh allowed to calculate sugar content in berries ($\gram$ Sugar Berry$^{-1}$).
%Maturity is estimated through the ratio between Sugar Concentration and acidity in grapes. 
 
 \subsection{Formalizing knowledge} \label{sec:K}

In this section, our aim is to show how ontologies can be used to formalize domain knowledge and to design a software sensor. 

In information science, an ontology formally represents knowledge as a set of concepts within a domain, and the relationships between pairs of concepts. 

Ontologies are becoming increasingly popular, due to the great amount of available (complex) data and to the need for modeling (qualitative) knowledge and structural information. This need first arose out of the development of the World Wide Web. However, there are still very few attempts to combine ontologies and statistical or data-driven models. This could be particularly useful in Life Sciences  and Agronomy \cite{Villanueva08, Thomopoulos13,destercke2013}.

The main incentives for using ontologies \cite{Gaurinoal09} are the following ones: 
\begin{enumerate} [1.]
\item To share a common understanding of structured information \cite{musen1992dks};
\item To explicit the specificities of domain knowledge;
\item To identify ambiguous or inappropriate model choices.
\end{enumerate}

For the present work, a specfic ontology has been built, in order to formalize the concepts and relations required to design a vine water deficit indicator and to analyze its impact on grape quality.

\begin{figure}[hbtp]
\begin{center}
\includegraphics[width=1.1\textwidth]{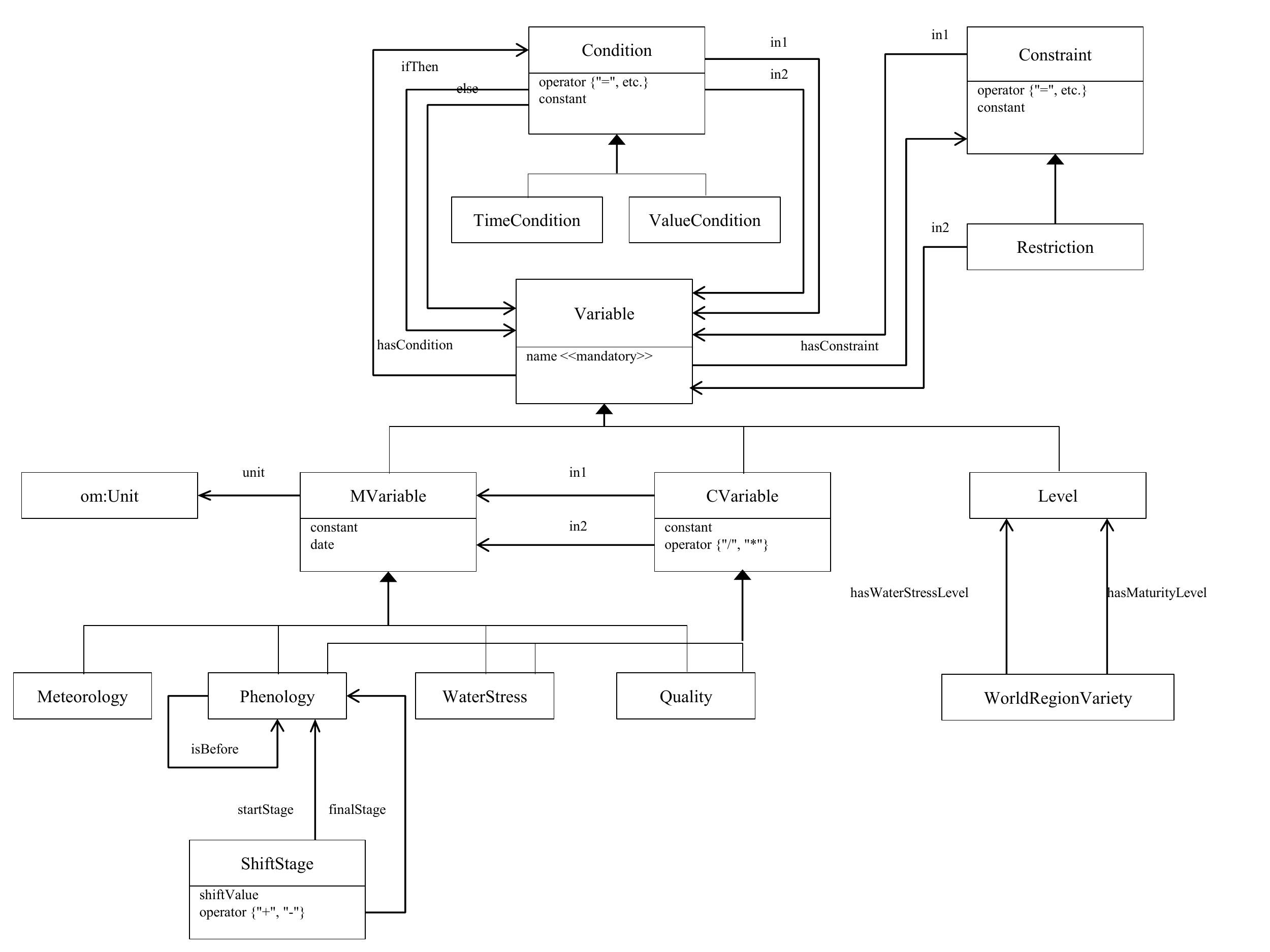}
\caption{Class diagram of the ontology  of Vine Water Stress }
\label{fig:ontoOVWS}
\end{center}
\end{figure}

The general class diagram of the ontology, called Ontology of Vine Water Stress (OVWS),  is shown on Figure \ref{fig:ontoOVWS} as a Unified Model Language (UML)  diagram. It is composed of concepts, represented as rectangular boxes, and of relations, represented by arrows. Formally, the ontology $\Omega$ is defined as a tuple $\Omega = \lbrace \mathcal{C} , \mathcal{R} \rbrace$ where  $\mathcal{C}$ is a set of concepts and $\mathcal{R}$ is a set of relations. 

Let us comment the main concepts and relations.

\subsubsection{Concepts}\label{sec:con}
In this ontology, four kinds of primary concepts were defined: \textit{Variable}, \textit{Condition}, \textit{Constraint} and \textit{ShiftStage}. All other concepts are subconcepts of these primary ones and linked to them by a \textit{subsumption} relation, as explained in Section \ref{sec:rel}. For instance, in Figure \ref{fig:ontoOVWS}, $Meteorology$, $Phenology$, \textit{WaterStress}  and $Quality$ are subconcepts of \textit{Variable}.
\begin{itemize} 
\item All \textit{variables} must have a name, they can have a date, a unit and a default value. The units are taken from $\mathit{OM}$, an ontology of units of measures and related concepts \cite{rijgersberg2013}.
.
\item The \textit{Condition} concept is defined with a comparison operator and two operands.  It will be used together with the \textit{hasCondition} ($\hascond_c$) relation, defined in Section \ref{sec:rel}.

\item The \textit{Constraint}  concept is defined with a comparison operator and one operand.  It will be used together with the \textit{hasConstraint} ($\hasconstr_c$) relation, defined in Section \ref{sec:rel}. The \textit{Restriction} concept is a subconcept of \textit{Constraint}, and is a specific two-fold constraint.

\item The \textit{ShiftStage} concept is proposed in order to determine a phenological stage from another one. This is the case for the $Nouaison$ stage, which is not generally observed. Its date can be estimated  by shifting the $Bloom$ date by $k$ $GDD$, where $k$ can be variety-dependent. $Nouaison$ and $Bloom$ are instances of the $Phenology$ concept.
\end{itemize} 
\subsubsection{Relations}\label{sec:rel}
On Fig. \ref{fig:ontoOVWS}, there are two kinds of arrows:  regular arrows and thick-headed ones. The former correspond to the \textit{subsumption} relation, and the latter to the other relations. In that last case, the arrow label gives the relation name, for instance \textit{hasCondition}.

\begin{itemize} 
\item The \textit{subsumption} relation, also called the `kind of' relation and denoted by $\isa$,  defines a partial order over $\mathcal{C}$. Given a concept $c \in \conceptset$, we denote by $\conceptset_c$ the set of sub-concepts of $c$, such that:
\begin{equation}
\conceptset_c=\condset{ c' \in \conceptset} { c'\isa c }.
\label{eqnCc}
\end{equation}
For example, in Figure~\ref{fig:ontoOVWS}, let us consider the concept $c=\textit{Variable}$. We have $\conceptset_{\text{Variable}}=\{ \textit{MVariable}, \textit{CVariable}, \textit{Level} \}$, where \textit{MVariable} represents a measurement available in a data base, \textit{CVariable} a variable calculated following a given method, and \textit{Level} a constant value depending on some other concepts.

\item The \textit{subsumption} relation can be multiple.
For instance a $Phenological$ $concept$ can be such as $c \isa CVariable$ or $c \isa MVariable$.

\item The $\mathit{isBefore}$ relation allows to represent temporal precedency. It is very important for checking the consistency of the phenological stage dates, where $bloom$ has to occur before $veraison$, and so on.

\item The  HasCondition ($\hascond_c$) relation, where $c$ represents the concept on which the condition is to be applied, is used together with a condition. 

%For instance,  $t < t_{max}$ used in the hypotheses of Section \ref{sec:kc} will be implemented using the $K_c \isa CVariable$ concept, the  $\hascond_c$ relation and the $Condition$ with a comparison operator $\le$.

\item Similarly, the HasConstraint ($\hasconstr_c$) relation allows the application of a $constraint$ on the $c$ concept. 

%All of the selection rules listed in Section \ref{sec:kc} can be implemented using $constraints$, or $restrictions$.

\end{itemize}

Examples of use will be given in Section \ref{sec:kcb}.

The ontology is modelled using the Web Ontology Language (OWL). OWL is a semantic markup language for publishing and sharing ontologies on the World Wide Web, which is specified using W3C\footnote{http://www.w3.org/TR/} recommendations. The use of OWL allows reusing ontologies developed elsewhere, such as \textit{OM}.

Note that each modeling component: data base, ontology and mathematical modelling or data analysis, can be modelled independently, using its favorite language (sql, R, OWL). The communication between the various components can be implemented by a high level interface, written in Python, PHP or Java.

\subsection{Design of the software sensor for vine water stress estimation}\label{sec:calibSapFlow}
Based on the knowledge formalized in the ontology given in Fig.\ref{fig:ontoOVWS} and on a mathematical model \cite{Ferreira2012}, the software sensor is designed to transform the calibrated transpiration measurements from sap flow sensors into a vine water deficit estimator, denoted by $\mathit{Ks(t)}$.

\begin{center}
\begin{figure}[htbp]
\centering\includegraphics[width=0.99\linewidth]{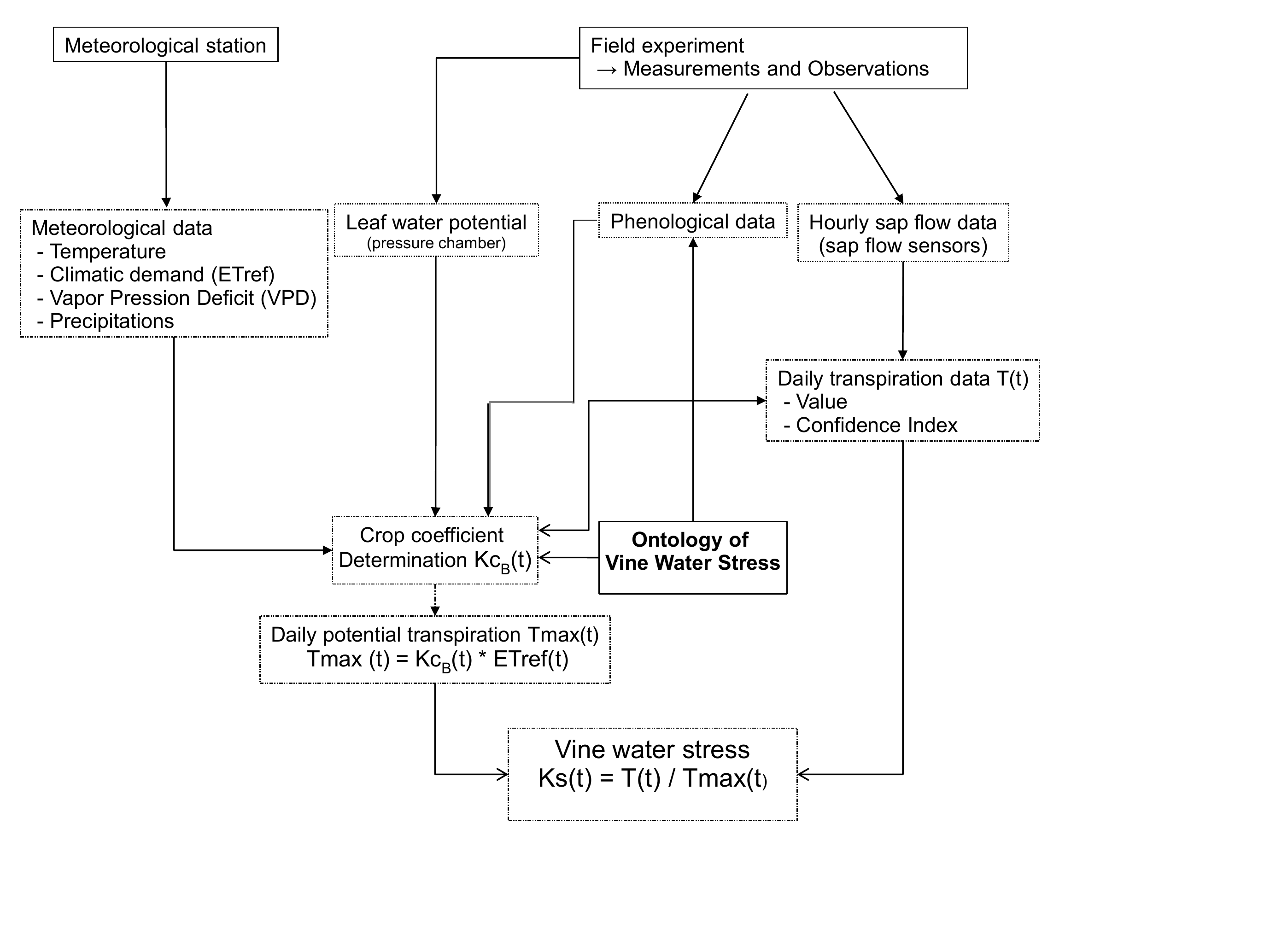}
\caption{Methodological steps for the vine water deficit ($\mathit{Ks}$) estimation.}
\label{fig:stressm}
\end{figure}
\end{center}

The different steps of the methodology used to design $\mathit{Ks}$ are summarized in Fig.\ref{fig:stressm}  and detailed in the following subsections.

\subsubsection{Sap flow under limiting soil water condition : computation of  $\mathit{Ks}$}\label{sec:ks}

%We used the approach described in \cite{Ferreira2012}. When a moderate or intense water stress is imposed,  a stress coefficient ($\mathit{Ks}$) has to be estimated. 

$\mathit{Ks}$ is the ratio between actual and maximum crop transpiration, defined as:
\begin{equation}\label{eqn:ks}
\mathit{Ks(t)}= \frac{T(t)}{\mathit{T_{max}(t)}}
\end{equation}
It accounts for the decline in vine water use due to soil moisture deficit. %The relationship between $\mathit{Ks}$  and predawn LWP has been shown \cite{Ferreira2012}. 
Even if some authors \cite{Allen1998} present a general proposal for estimating $\mathit{Ks}$, specific $\mathit{Ks}$ functions for vineyards, from field experiments, are not generally available.

In the vine context, in Eq.\ref{eqn:ks}, $T$ is the daily measured transpiration from sap flow and $\mathit{T_{max}}$ is the daily maximal vine transpiration obtained under dry soil condition (meaning no cover crop) when soil moisture is non limiting, defined as in \cite{Allen1998}.
\begin{equation}\label{eqn:Tmax}
T_{max}(t) = {Kc_B}(t) \ {ET}_{ref}(t)
\end{equation}

$\mathit{ET_{ref}}$ is the reference evapotranspiration and $\mathit{Kc_B}$ a coefficient linearly related to the leaf area index (LAI) or to the fraction of ground coverage \cite{Williams2005, Picon-Toro2012}. As $\mathit{Kc_B}$ is dependent upon crop type and management practices, which will influence the rate of canopy development and the ultimate canopy size, \textit{i.e.} amount of ground cover (\cite{Allen1998}), a site-specific determination of $\mathit{Kc_B}$ is necessary  for each vineyard.

\subsubsection{Sap flow under non limiting soil water condition : computation of dry soil $Kc_{B}$}\label{sec:kcb}
\begin{figure}[htbp]
%\centering\includegraphics[width=0.7\linewidth]{FigTheorKc2.png}
\centering\includegraphics[width=0.7\linewidth]{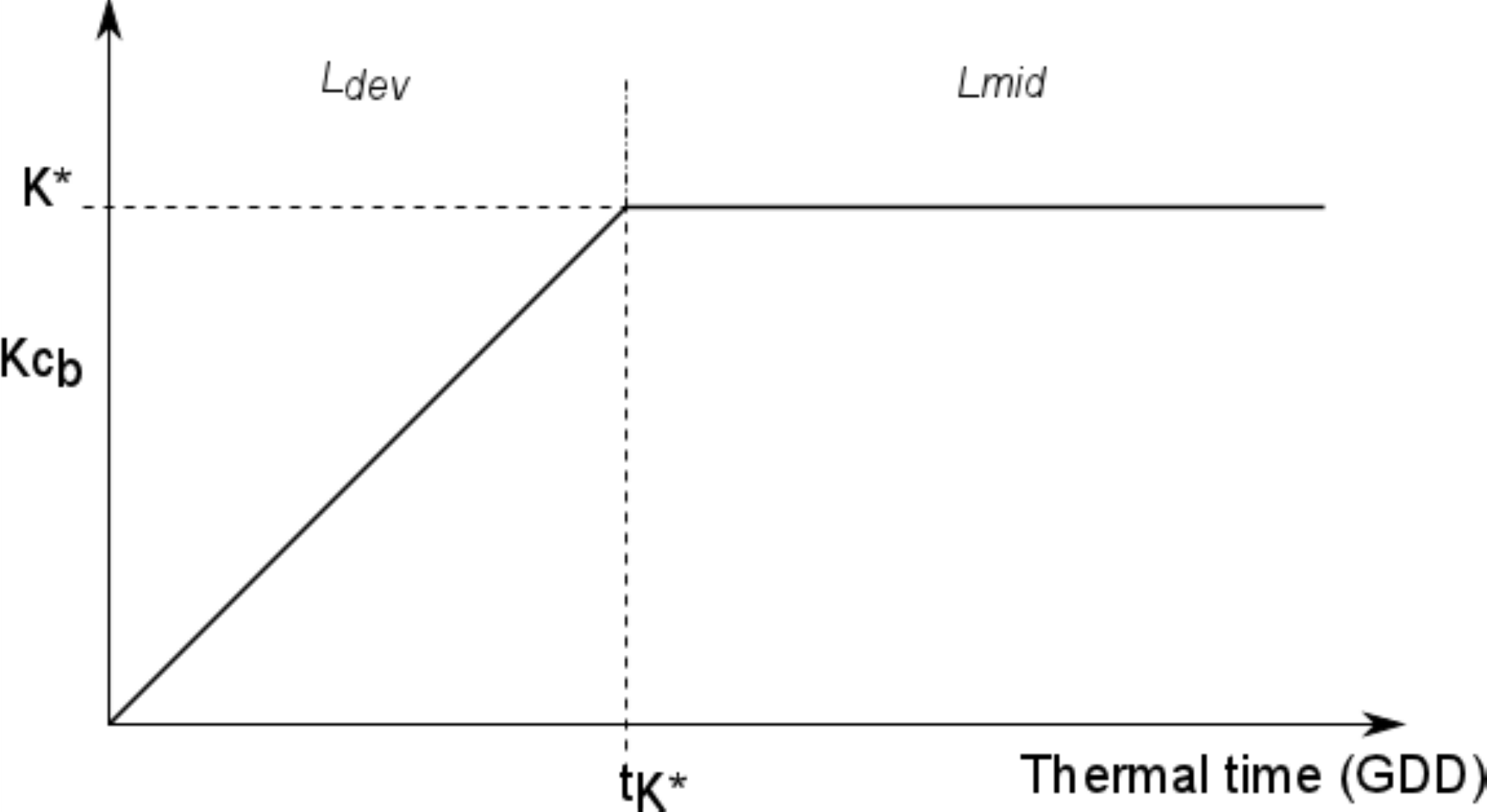}
\caption{Theoretical curve of $Kc_B$ evolution during the season.}
\label{fig:Kc}
\end{figure}

No direct measurements are available to determine $\mathit{Kc_B}(t)$ (see Eq.\ref{eqn:Tmax}), which is vine specific and varies with leaf area. We propose to use formalized concepts and relations based on expertise, all of them implemented in the OVWS ontology. $\mathit{Kc_B}$ profile is divided into two growth stages : $\mathit{L_{dev}}$ and $\mathit{L_{mid}}$ as presented in Fig.\ref{fig:Kc}. To determine $Kc_B(t)$, two hypotheses on the curve shape are assumed:
\begin{eqnarray}\label{eqn:K*}
\mathit{Kc_B}(t)=f(t) \mbox{ for } t< t_{K^*}\\
\mathit{Kc_B}(t)=K^* \mbox{ for } t\ge t_{K^*}
\end{eqnarray}
where $f(t)$ is assumed to be linear in $t$, and $t_{K^*}$ is the breakpoint for which $Kc_B$ reaches the plateau $K^* $.
The key point is to set $t_{K^*}$, or indifferently $\mathit{K^*}$.
According to Eq.\ref{eqn:Tmax}, we make the hypothesis that, in the absence of water deficit, $\mathit{K^*}$ is defined as:

\begin{equation}\label{eqn:kc}
\mathit{K^*} = \frac{\mathit{T}(t_{K^*})}{\mathit{ET}_{ref}(t_{K^*})}
\end{equation}

Using the OVWS ontology defined in Section \ref{sec:K}, the following rules are set up to  automatically define a limited number of potential options for $t_{K^*}$. The interest of having the rules and concepts defined in an ontology is two-fold: $i)$they have to be completely explicit,  $ii)$they can evolve independently of the numerical procedures.
\begin{enumerate}
\item{\textit{Selection based on phenology}}\label{para:pheno}

A linear relationship exists between $\mathit{Kc_B}$ variations and leaf area index (LAI)  or the fraction of ground covered by the vine \cite{Ferreira2012}. We thus assume that peak $\mathit{Kc_B}$ (i.e. $\mathit{K^*}$) is reached when LAI stops increasing. Consequently, the search period for $\mathit{K^*}$ has been limited to the period between budbreak and veraison.

These two concepts are defined as subconcepts of the Phenology concept, itself being a subconcept of \textit{MVariable}. The period limitation is intanciated by two \textit{TimeConditions}, applied onto the $\mathit{Kc_B} \isa CVariable$ concept, the  $\hascond_c$ relation where the $Condition$ is characterized by a comparison operator $\le$ (resp. $\ge$) and the \textit{Veraison} (resp. \textit{Budbreak}) concept.

\item{\textit{Selection based on predawn leaf water potential}}\label{para:psyb}

Conditions of maximal soil moisture availability could be inferred from predawn leaf water potential measurements, associated with a confidence interval derived from VPD. A rule was set so that $\mathit{K^*}$ has to be reached before the first day at which predawn LWP measurement reveals a water stress. The levels to which predawn LWP characterises water stress can be defined by the stakeholder, or else set in agreement with a standard level, based on a region or/and variety. \\
This is implemented in the ontology by the \textit{WorldregionVariety} and \textit{level} $\isa Variable$ concepts.

\item{\textit{Selection based on meteorology}}\label{para:meteo}

Transpiration measurement through sap flow is sensitive to climatic conditions, mainly light and $\mathit{VPD}$. To account for sensitivity of transpiration measurements to $\mathit{VPD}(t)$, a filtering rule was set to remove computed $Kc_B(t)$ obtained in situations of heat spikes, defined as period with $\mathit{VPD}$ greater than a given level, set to 3.5 $\kilo\pascal$ in the present case study. \\
The rule is implemented using a $\hasconstr_c$ relation, applied onto the \\$\mathit{Kc_B} \isa CVariable$ concept, where the $Constraint$ is characterized by a comparison operator $\le$  and the $\mathit{VPD} \isa Meteorology \isa MVariable$  concept.

\item{\textit{Selection based on curve shape}}\label{para:deriv}

By definition, $\mathit{K^*}$ is reached when the ratio $\displaystyle \frac{\mathit{T}(t)}{\mathit{ET_{ref}(t)}}$ reaches a maximum during a few days (as $T(t)$=$T_{max}(t)$ and $Kc_B(t)$=$K^*$) and then decreases (as $T(t)$<$T_{max}(t)$ due to limiting soil water conditions while $Kc_B(t)$=$K^*$). As such, potential options for $t_{K^*}$ have been defined at points with a null first derivative and a negative second derivative. \\
This selection is implemented using two concepts: $\dot{\mathit{K^*}}$ and  $\ddot{\mathit{K^*}}$, both such as $\isa WaterStress \isa CVariable$ , and a $\hasconstr_c$ relation with a \\$Constraint$  characterized by a comparison operator $\le \epsilon$ or $\le 0$. 

\end{enumerate}
\paragraph{\textit{User selection based on expert knowledge}}\label{para:choiceKc}
\

The analysis of $\displaystyle \frac{\mathit{T}(t)}{\mathit{ET}_{ref}(t)}$ curve shape, associated with all previous rules based on phenology, meteorology and predawn leaf water potential leads to the proposal of a small finite set of  $\mathit{t_{K^*}}$ candidates.

The final choice is left to the stakeholder who is the best aware of the management practices or particular uncontrolled events that could have interfered with vine growth (irrigation, leaf removal, treillis system, ...) and therefore $\mathit{Kc_B}$ curve.

%%%%%%%%%%%%%%%%%%%%%%%%%%%%%%%%%%%%%%%%%%%%%%%%%%%%%%%%%%%%%%%%%%%%%%%%

\subsection{Relating software sensor output to product quality} \label{sec:stat}
The software sensor output consists of temporal data, and various methods can be used to study the relationships between these data and product quality. Two complementary lines of work, based on statistical methods, are explored in the present work.  The first one consists of extracting significant  scalar parameters from the temporal data and using them as input to decision trees, in order to provide the most discriminant features. The second one uses functional data analysis, that gives the possibility to model the temporal data impact on product quality as a whole. However, curve analysis is a recent research topic, with relatively few methods available, in comparison with classical data analyses.

This section is divided into three parts. The first part describes how to use the formalized knowledge for extracting significant  scalar parameters from the temporal data. The other two parts give some elements necessary to understand the statistical methods that will be used: decision trees and functional analysis.

\subsubsection{Extracting   scalar parameters from  software sensor output}\label{sec:SaggrMM}

Meaningful scalar parameters can be extracted from temporal courses determined by the software sensor outputs. In many cases, expert knowledge can be the support of such extraction procedures. In the case of vine water courses, this can be achieved by taking into account important phenological periods, which are defined as concepts in the ontology (see Fig.\ref{fig:ontoOVWS}).

Three periods were first defined according to phenological stages: the whole season, the pre-veraison period, which goes from the nouaison stage to the veraison stage, and the post-veraison period which ranges from the veraison stage to the harvest date. In a second step, the post-veraison period was divided by taking into account the maturity stage of berries (see Section  \ref{sec:pheno}), which allowed to a add a fourth period ranging from veraison to maturity. Maturity stage is reached when the ratio between Sugar Concentration and acidity in grapes yields a given threshold, defined according to variety

Using trapeze integration under $\mathit{Ks}$ curves over these four periods, the continuous  $\mathit{Ks(t)}$ curve was summarized into four new variables corresponding to the cumulative amount of stress encountered by the vine over these periods: \textit{NouHarv}, \textit{NouVer}, \textit{VerHarv} and \textit{VerMat}. Table \ref{tab:scalar} gives the summary of these four aggregated variables for each plot. Since all these aggregated variables are based on the area under the curve, the lower their value, the  stronger the water stress over the considered period. 

\begin{table}[hbtp] 
\begin{tabular}{|l|l|l|l|l|l|l|}
\hline
Site&Variety&Irrigation& NouHarv&    NouVer&VerHarv&VerMat\\
\hline\hline
LB-CS&Cabernet-S.&$i_0$&1165.6&704.6&432.6&432.6\\
&&$i_1$&1117.8&713.8&375.0&347.6\\
\hline
OUV-Mer&Merlot&$i_0$&742.1&457.2&278.3&163.9\\
&&$i_1$&1233.8&814.7&403.4&247.9\\
\hline
StGER-Mer&Merlot&$i_0$&608.9&470.9&131.6&129.4\\
&&$i_1$&808.3&473.4&327.0&214.5\\
\hline
PR-Mer&Merlot&$i_0$&655.4&381.8&250.2&199.5\\
&&$i_1$&722.7&398.3&313.4&266.2\\
\hline
StSAU-Char&Chardonnay&$i_0$&695.3&442.6&241.9&213.6\\
&&$i_1$&693.8&414.0&265.3&253.7\\
\hline
RIE-Gre-Chm&Grenache&$i_0$&651.1&465.2&169.9&169.9\\
&&$i_1$&620.0&390.9&209.5&209.5\\
\hline
RIE-Gre-Chp&Grenache&$i_0$&512.3&380.0&123.0&NA\\
&&$i_1$&963.9&580.2&362.0&362.0\\
\hline
PIO-Gre&Grenache&$i_0$&677.2&458.4&212.0&152.3\\
&&$i_1$&887.4&514.4&363.8&180.8\\
\hline

\end{tabular}
\centering\caption{Values of aggregated variables for each site-variety-irrigation treatment combination (i) over the entire season \textit{NouHarv}, (ii) before veraison \textit{NouVer}, (iii) after veraison \textit{VerHarv} and (iv) from veraison to maturity \textit{VerMat}.}
\label{tab:scalar}\end{table}

\subsubsection{Decision trees as interpretable models \label{sec:dt}}
Decision tree algorithms are well established learning methods in supervised data mining and statistical multivariate analysis. They allow to display non linear relationships between features and their impact on a response variable, in a compact way. A low complexity  \cite{BenDavid06}  is essential for the model to be interpretable, as confirmed by Miller's conclusions
\cite{Miller56} relative to the \textit{magical number} seven. 

Decision trees can handle classification problems or regression cases, depending on the nature of the response variable.
We present here the regression case, where the response variable is numerical.

Input to regression decision trees consists of a collection of $N$ training cases, each having a tuple of values for a set of $P$ input variables, and one output variable  $(\mathbf{x_i, y_i}) = (x_{1,i}, x_{2,i} \ldots x_{P,i}, y_i).$
An input $X_p$ ($p=1...P$) is continuous or discrete and takes its values $(x_{p,i})_{i=1\dots N}$ on a domain $\mathcal{X}_p$.  The goal is to learn from the training cases a recursive structure (taking the shape of a rooted tree) consisting of (i) leaf nodes labeled with a mean value and a standard deviation, and (ii) test nodes (each one associated to a given variable) that can have two or more outcomes, each of these linked to a subtree.

%On a given node, the algorithm examines in turn all available variables, and selects the variable that most effectively splits the set of samples into subsets improving the separation between output values. Once (and if) a variable is selected, a new test node is created that splits on this variable, and the procedure is recursively applied on each (new) node child. At each node, the algorithm stops when no more variables are available, or if there is no improvement by splitting further: the node then becomes a leaf. 

Decision trees are easily interpretable for a non-expert in statistical or learning methods, and facilitate exchanges with the domain expert.

Well-known drawbacks of decision trees are the sensitivity to outliers and the risk of overfitting.
To avoid overfitting, cross-validation is included in the procedure and to gain in robustness, a pruning step usually follows the tree growing step (see~\cite{quinlan1986idt,breiman1car,quinlan1993cpm}).

Note that the CART family~\cite{breiman1car}, based on binary splits. is mostly used by statisticians. There is another tree family~\cite{quinlan1986idt}, called ID3 \cite{quinlan1993cpm}, allowing non binary splits and mostly used by artificial intelligence researchers. 

In this work, we used CART-based trees. In that case, the splitting criterion is based on finding the one predictor variable (and a given threshold of that variable) that results in the greatest change in explained deviance (for Gaussian error, this is equivalent to maximizing the between-group sum of squares, as in an ANOVA).  This is done using an exhaustive search of all possible threshold values for each predictor. The implementation used for decision trees is the \cite{R} software with the \textit{rpart} package\footnote{http://cran.r-project.org/web/packages/rpart/index.html}. Specifying variety, \textit{NouVer}, \textit{VerHarv} and \textit{VerMat} as explanatory variables, we computed decision trees on maximum values of grape quality features over the season.

%\color{red}{ \bigskip remarque bch: plot classif - enleve de l'article stat-info }
%

%\subsubsection{Plot classification}
%
%Unsupervised classification methods allow partitioning a set of \textit{n} observations into \textit{k} homogeneous groups (\textit{k}<<\textit{n}) according to \textit{p} explanatory variables. Unsupervised classification methods can be devided into two main kinds: classifications based on distance and classifications based on k-means. The former use a measure of dissimilarity between the \textit{n} observations associated with a linkage criterion  to decide which observations should be combined in clusters. The k-means algorithm is a method of cluster analysis in which each observation belongs to the cluster with the nearest mean, the number of clusters \textit{k} being set previously.
%Using the cumulative amount of stress encountered by the vine over the two main periods of the season (i.e. pre-veraison (\textit{NouVer}) and post-veraison (\textit{VerHarv}) periods), we performed a unsupervised classification based on the euclidean distance and the Ward linkage method to partition the site-variety-irrigation treatment combinations into homogeneous groups. This first partition allowed to choose an optimal number of $k$ clusters. In a second step, a k-means clustering with $k$ cluster was performed to give an explanation of each cluster according to the two variables \textit{NouVer} and \textit{VerHarv}. 
\color{black}
\subsubsection{Functional data analysis\label{sec:fdata}}
Functional linear regression is an approach to model the relationship between a scalar dependent variable $Y$ and a functional predictor $X(t)$, a function of a real variable $t$ (time for example). The model is written as
\begin{equation} \label{model-FLR}
Y_i = \beta_0 + \int X_i(t) \beta(t) dt + \varepsilon_i \, , \quad i=1,\dots,n
\end{equation}
where $\varepsilon_i$ is a random error, $\beta_0$ is the intercept of the model and $\beta(t)$ is the coefficient function, both unknown and to be estimated from independent observations $(X_i(t),Y_i)_{i=1,\dots,n} $. In this model, $\beta(t)$ determines the effect of $X_i(t)$ on $Y_i$. For example, $X_i(t)$ has a greater effect on $Y_i$ over regions of $t$ where $|\beta(t)|$ is large. On the opposite, $X_i(t)$ has no effect on $Y_i$ over regions of $t$ where $\beta(t)$ is zero. Estimating $\beta(t)$ in model (\ref{model-FLR}) has given rise to an increasing litterature during this last decade, see for example~\cite{RS05}. A common approach involves projecting $\beta$ and the $X_i$'s in a $p$-dimensional basis function where $p$ is large enough to capture the unknown variations of $\beta$, but small enough to regularize the fit. 

Interpretation of such estimators is not that easy. Recently, \cite{james2009} introduced new estimators that are both interpretable, flexible and accurate. The method, called ``Functional Linear Regression That's Interpretable'' (FLRTI), is based on a particular basis function and model selection principles.
The $\beta$ function is either assumed to be zero or to have a simple linear form. The reason behind the first assumption is that the $X_i(t)$ observations are not of equal importance to explain the response $Y_i$. As said above, $X_i(t)$ has no effect on $Y_i$ when $\beta(t)=0$. The second assumption is made for technical reasons to simplify the model. All together, this model will have a high explanatory capability, contrary to a pure predictive model.
These assumptions will constraint the estimation of the regression model (\ref{model-FLR}), which corresponds to a penalized regression in sparse models. Two tuning parameters have to be fixed, a penalty term $\sigma$ and a weight $\omega$. The penalty term is part of the model selection process (by lasso or Dantzig selector). The largest the $\sigma$, the more the form-related constraint is enforced. The weight $\omega$ impacts the relative number of zeros of the $\beta$ function. A weight of 0 indicates that only the linear form constraint is respected. 

The FLRTI method is implemented in an R function available on J. Gareths's web page\footnote{http://www-bcf.usc.edu/~gareth/research/flrti}. A cross-validation algorithm is also proposed to optimize the choice of $\sigma$ and $\omega$.

Using the FLRTI method, we analysed the effects of water stress curve over the season ($\mathit{Ks(t)}$) on Berry Weight and Berry Sugar Concentration at harvest. %To define the $\lambda$ and weight values, we ran several models using combinations of weight (from 0 to 1 by 0.05) and sigma (from 0.00001 to 0.5) and choosed the best model according to cross validation errors. We ended up with a model with a weight of 0.05 and a sigma from 0.01. 

%%%%%%%%%%%%%%%%%%%%%%%%%%%%%%%%%%%%%%%%%%%%%%%%%%%%%%%%%%%%%%%%%%%%%%%%

\section{Results} \label{sec:resu}
In this section, we first present the results of the $\mathit{Ks(t)}$ estimation using the software sensor. In a second step, we study the relationship between $\mathit{Ks(t)}$ and grape quality features, using the methods described in Section \ref{sec:meth}. In the following, we will refer to irrigated treatments with $i_1$, and to non irrigated ones with $i_0$.

\subsection{Vine water stress course $\mathit{Ks(t)}$ estimation}\label{sec:resKs}
Sap flow data require a pre-treatment, including sensor selection and signal smoothing. Sap flow sensors have only been used recently in European vineyards. Thus, calibration protocols are not established yet and therefore sensors can still be unreliable. Consequently, a selection step is required.

Sensor reliability has been assessed on the basis of the number of incorrect hourly measurements resulting from expert filtering methods. A sensor was considered reliable when less than 5\% of the hourly data were filtered out. For each variety-irrigation combination, the mean daily vine transpiration was calculated as the mean of daily measures from reliable sensors, which helped limiting the variability in plant transpiration measurements.
However, one of the major drawback of sensor selection was the potential lack of reliable measurements on a daily basis.

To capture important patterns in daily sap flow data, while leaving out noise and extreme variations (daily peak), sap flow courses were smoothed with the central moving average method with a five day window. This smoothing allowed the removal of missing values and extreme peaks.

%Both procedures are detailed in Appendix A.%\ref{sec:sensorsel}

\subsubsection{$K^*$ determination} \label{sec:resKc}
	
\begin{figure}[htbp]
\centering\includegraphics[width=0.99\linewidth]{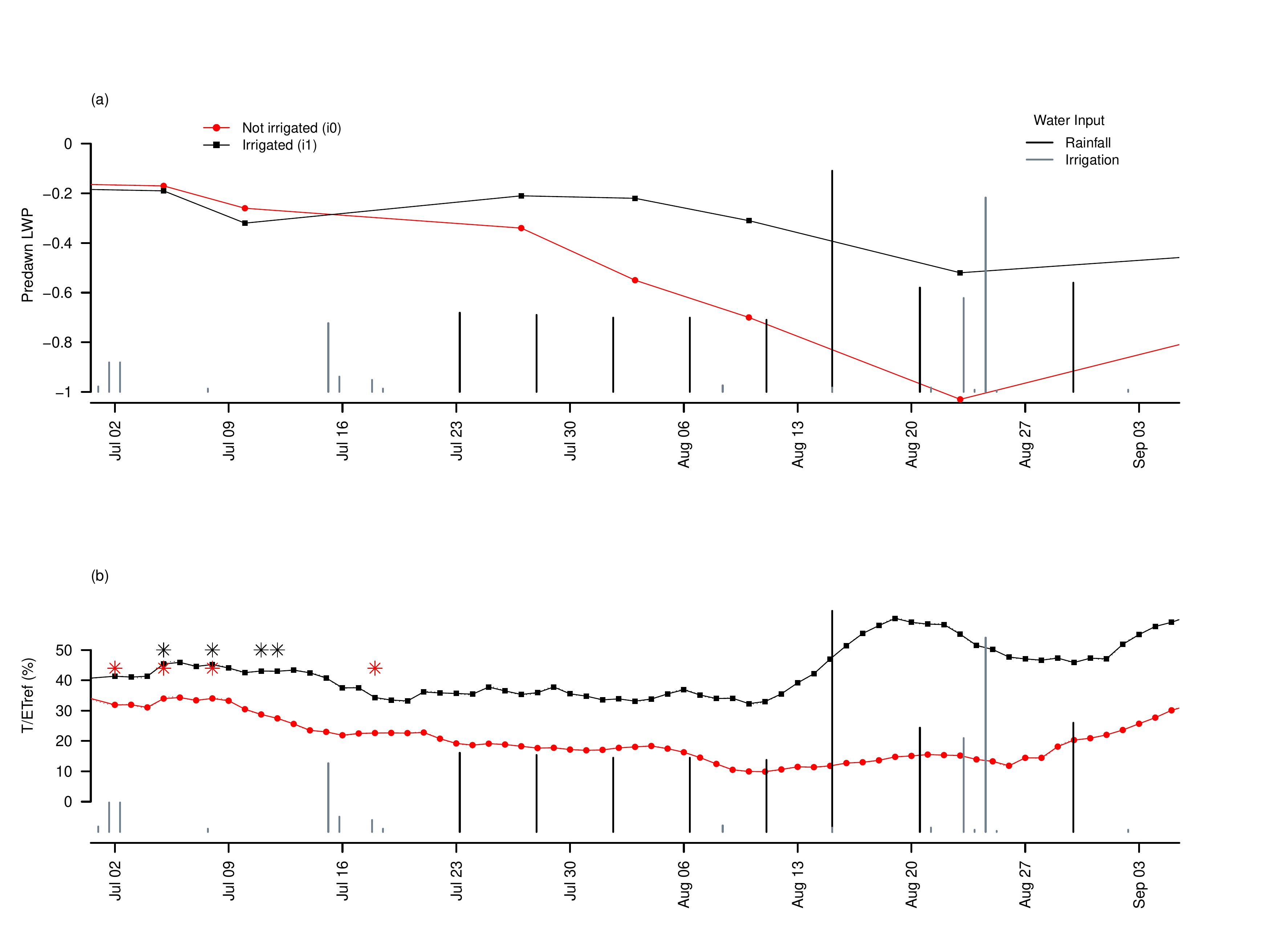}
\caption{(a) Vine water status as indicated by leaf water potential and (b) determination of $K^*$ and $t_{K^*}$ (stars) based on expert knowledge following the  mathematical algorithm suggestions for non irrigated (red bullets) and irrigated (black squares) treatments on the Grenache variey at Piolenc site. }
\label{fig:Kcdet}
\end{figure}

Regarding all site-variety combinations, the knowledge-based algorithm for $t_{K^*}$ determination proposed from 5 to 9 candidates (resp. from 4 to 8 candidates) in the non-irrigated $i_0$ (resp. irrigated $i_1$) treatments. Most of the dates proposed by the mathematical algorithm were in accordance with expert knowledge, so allowing the expert to choose $t_{K^*}$ within the algorithm suggestions (Fig.\ref{fig:Kcdet}). The results are given in Table \ref{tab:kc}.

\begin{table}[hbtp] 
\begin{center}
\begin{tabular}{l|l|l|l|l|l}
Site & Variety & Irrigation & $K^*$ & $t_{K^*}$ (GDD)&First irrigation(GDD)\\
\hline
La Baume & CS &i0 & 20.3 & 677.9 &\\
&&i1&32&698.6&1268.3\\
Pech Rouge & Merlot & i0 & 19.4 &614.5\\
&&i1&26,6&614.5&610.4\\
St Gervasy & Merlot &i0 &69.3 & 625.3&\\
&&i1&85.1&669.4&844\\
Ouveillan & Merlot &i0 & 37.1 &829.2&\\
&&i1&21&1005.1&939.7\\
Piolenc&Grenache&i0&44.3&530.1&\\
&&i1&58.1&530.1&864.4\\
Rieux&Grenache+&i0&43.4&600.5&\\
&&i1&29.1&594.5&789.5\\
Rieux&Grenache-&i0&29.2&580&\\
&&i1&46.1&580&789.5\\
St Sauveur&Chardonnay&i0&43.3&642\\
&&i1&54.4&749.7&777.2\\
\hline
\end{tabular}
\centering\caption{Values of basal crop coefficients $K^*$ and dates ($t_{K^*}$ in GDD) at which they were estimated in the site-variety-irrigation treatment combinations during season 2012.}
\label{tab:kc}
\end{center}
\end{table}

Fig.\ref{fig:Kcdet} illustrates the results for the Grenache variety at the Piolenc site.

\color{red}
%Note d'Aurelie Fig. 5: Je mettrais peut etre les traits d'irrigation et de precipitations un petit peu plus larges. J'ai peur que ca ne se voit pas tres bien si les figures sont reduites sur une colonne par exemple. 
%Dans le titre des graphiques mettre le B de KcB en indice.
%Attention, sur la figure du haut, on a rainfall et irrigation pour les water inputs, pour celle du bas, on a precipitations et irrigation. Homogeneiser : remplacer rainfall par precipitations
%Sur l'axe des y, mettre Predawn LWP sur la figure du haut (au lieu de pre-dawn LWP) et T/Etref sur celle du bas (au lieu de Tr/ETref car on a dit que la transpiration actuelle était notée T(t))

%\color{blue} bch : fait
\color{black}

The validity of the $K^*$ determination procedure can be assessed according to different points. 
First, the results regarding $K^*$ determination based on the coupling of mathematical algorithms and expert knowledge were consistent with existing literature. 
Indeed, most of $t_{K^*}$ occurred between 600 and 700 GDD after budbreak (Table \ref{tab:kc}), which is in accordance with $t_{K^*}$  reported in \cite{Picon-Toro2012} from a 3 year study in western Spain on Tempranillo, and in FAO-56 \cite{Allen1998}, that respectively reported  $t_{K^*}$ around 650 GDD and 555-592 GDD after budbreak. 

%It is also in agreement with the duration of growth stage suggested by FAO 56 \cite{Allen1998} yielding a $t_{K^*}$ around 555-592 in the growing conditions of a Merlot vineyard growing in a semi-arid climate. However, in a recent study Poblete-Echavarria et al, 2013 using sap flow sensors in a Merlot vineyard to characterize $K^*$ found that the developing of growth was shorter than than suggested by FAO 56. $t_{K^*}$ was reached on average around 517 GDD after bud break. Authors suggested that the shortening in the length of the growing stage could be attributed to climatic conditions at the study site.

\subsubsection{Maximal transpiration and $\mathit{Ks}$ estimation} \label{sec:TmaxKs}

\begin{figure}[htbp]
\centering\includegraphics[width=1\linewidth]{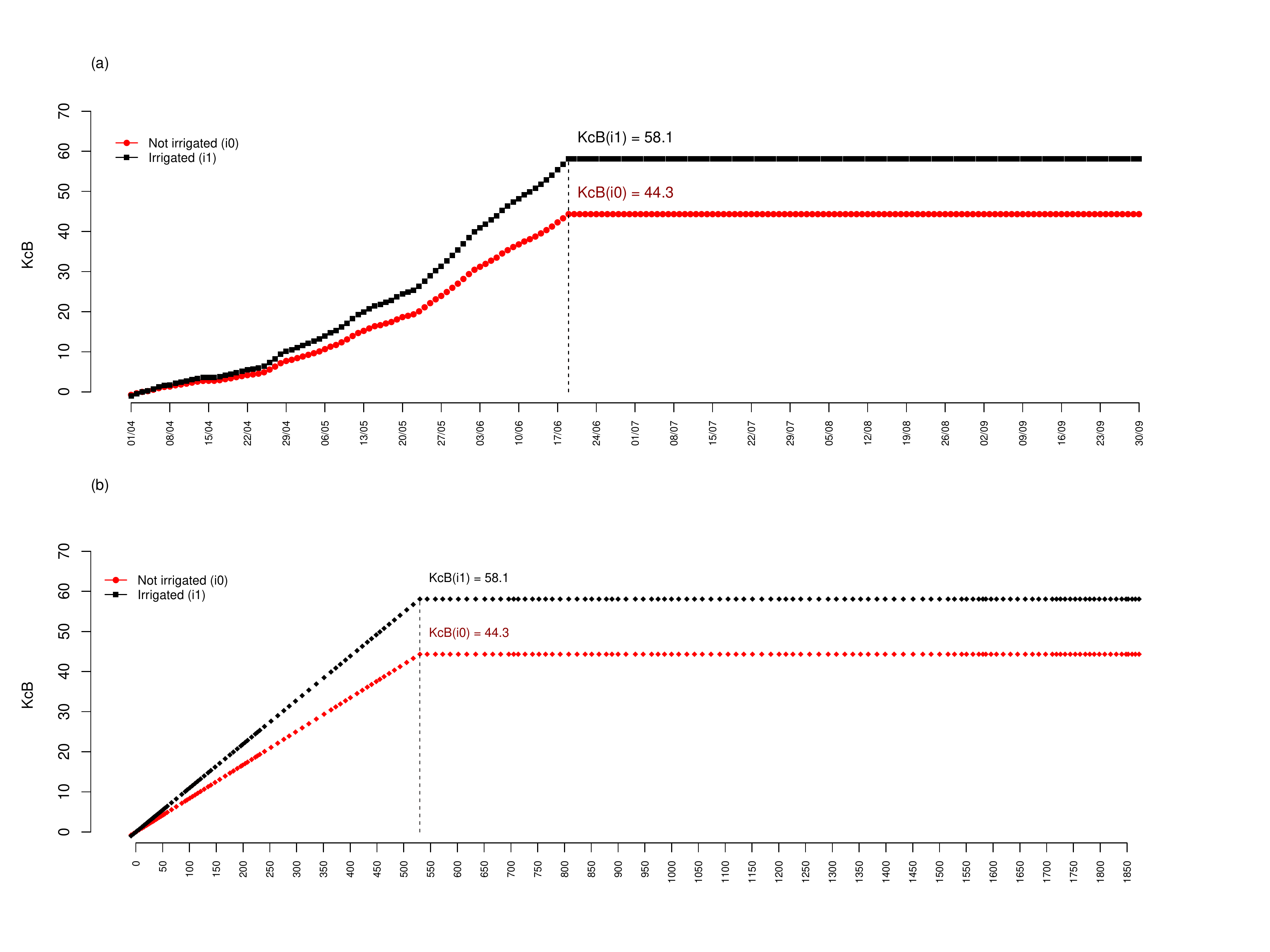}
\caption{Evolution of vine basal crop coefficient ($Kc_B$) during the season at Piolenc site with Grenache variety, from budbreak to harvest. (a) x-scale in Julian days - (b) x-scale in GDD since budbreak.}
\label{fig:kcCurve}
\end{figure}

Following determination of  $K^*$ and $t_{K^*}$, $Kc_B(t)$ was calculated over all $t$ values (Eq.\ref{eqn:K*}). 
Its evolution for a Grenache variety is plotted on Fig.\ref{fig:kcCurve}, both in calendar time (a) and thermal time since budbreak (b).

\color{red}
%Note d'Aurelie Fig 7: Il faudrait remonter un tout petit peu les labels afin qu'ils apparaissent mieux au dessus des courbes et eventuellement reduire la taille des points qui font la courbe. 
%Sur l'axe des y mettre KcB au lieu de Kc

%\color{blue} bch : fait
\color{black}

$Kc_B(t)$ was then used to calculate the daily vine maximal transpiration ($\mathit{T_{max}}$), according to Eq.\ref{eqn:Tmax}. Finally, $\mathit{Ks(t)}$ was calculated as the daily ratio of measured transpiration by reliable sensors over potential transpiration (Eq.\ref{eqn:ks}).

\begin{figure}[htbp]
\centering\includegraphics[width=1.2\linewidth]{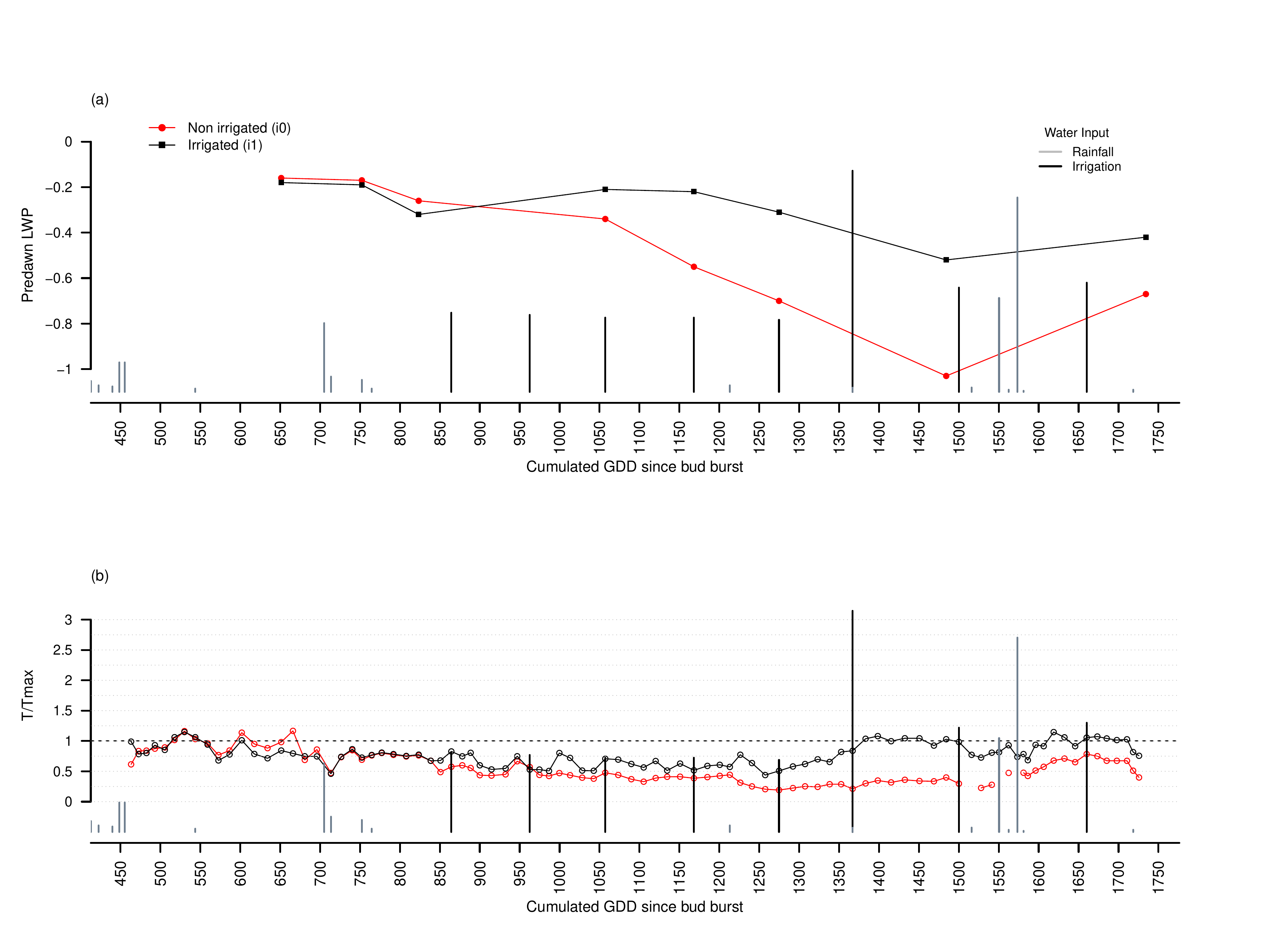}
\caption{Water stress during 2012 millesim in Grenache variety at Piolenc site assessed by (a) Predawn LWP and (b) vine water stress indicator $\mathit{Ks}$.}
\label{fig:Sstress}
\end{figure}

Figure \ref{fig:Sstress} shows vine water status according to both indicators: (a) Predawn LWP and  (b) $\mathit{Ks}$ during the season 2012 in a Grenache variety of the Languedoc-Roussillon region. 
%Based on  Eq. \ref{eqn:S}, when $\mathit{Ks}$< 1, vine's transpiration is lower than potential transpiration. We can assume that this is due to water deficit. Consequently, a declining value of $\mathit{Ks}$ over the season highlights that the vine is gradually experiencing more water deficit.

\color{red}
%Note d'Aurelie: same remark as for fig.5 + Attention sur l'axe y de la figure du bas. mettre T/Tmax (pour rester coherent avec l'equation 2 (au lieu de Tr/TrPot) 

%\color{blue} bch : fait
\color{black}

\subsection{Relationships between vine water stress $\mathit{Ks(t)}$ and grape quality}\label{sec:aroma}

As explained in Section \ref{sec:stat}, $\mathit{Ks(t)}$ can be used in two different ways, either summarized as a series of scalar values, or as a whole. The way to summarize $\mathit{Ks(t)}$ is detailed in Section \ref{sec:SaggrMM}. Scalar values and $\mathit{Ks(t)}$ will be put in relation to grape quality, by the respective use of (i) regression trees and (ii) functional data analysis. The studied grape quality features include {\it i)} Berry Weight and {\it ii)} Sugar Concentration in berries.

\subsubsection{Regression trees} \label{sec:rt}
Aggregated variables over periods can  be used as explanatory variables in regression trees to detect and prioritize the periods critical to changes in grape quality. We studied the effects of \textit{NouVer}, \textit{VerHarv}, \textit{VerMat} and variety on the two components of grape quality cited above. 

\begin{figure}[htbp]
\centering\includegraphics[width=0.9\linewidth]{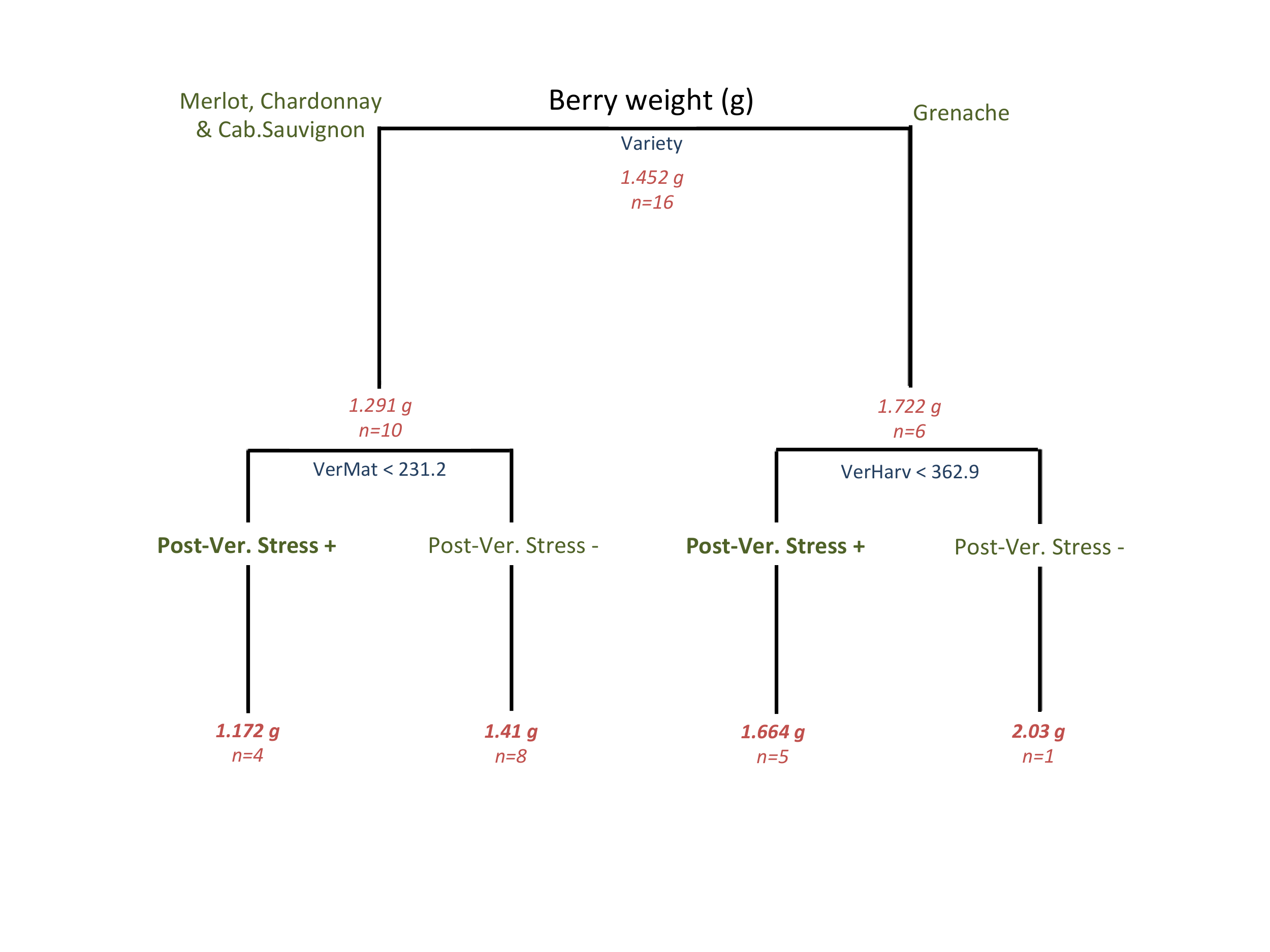}
\caption{Regression tree explaining Berry Weight ($\gram$) using scalars summarizing the three periods, i.e. pre-veraison (\textit{NouVer}), and post-veraison either until maturity (\textit{VerMat}) or harvest (\textit{VerHarv}).} %Cross validation error was 0.018.}
\label{fig:RTBW}
\end{figure}

According to Fig.\ref{fig:RTBW}, Berry Weight  seems to be mostly affected by the variety (Fig.\ref{fig:RTBW}). Grenache variety significantly yields heavier berries. The second split for all varieties is done on the post-veraison water stress only (either \textit{VerHarv} or \textit{VerMat}). The higher the post-veraison water stress, the smaller the Berry Weight.

%The increase in Berry Weight  as a consequence of higher water availability has been indeed widely reported \textcolor{red}{(e.g., Matthews and Anderson, 1988; Esteban et al., 1999, santesteban et al 2006)}.

\begin{figure}[htbp]
\centering\includegraphics[width=0.9\linewidth]{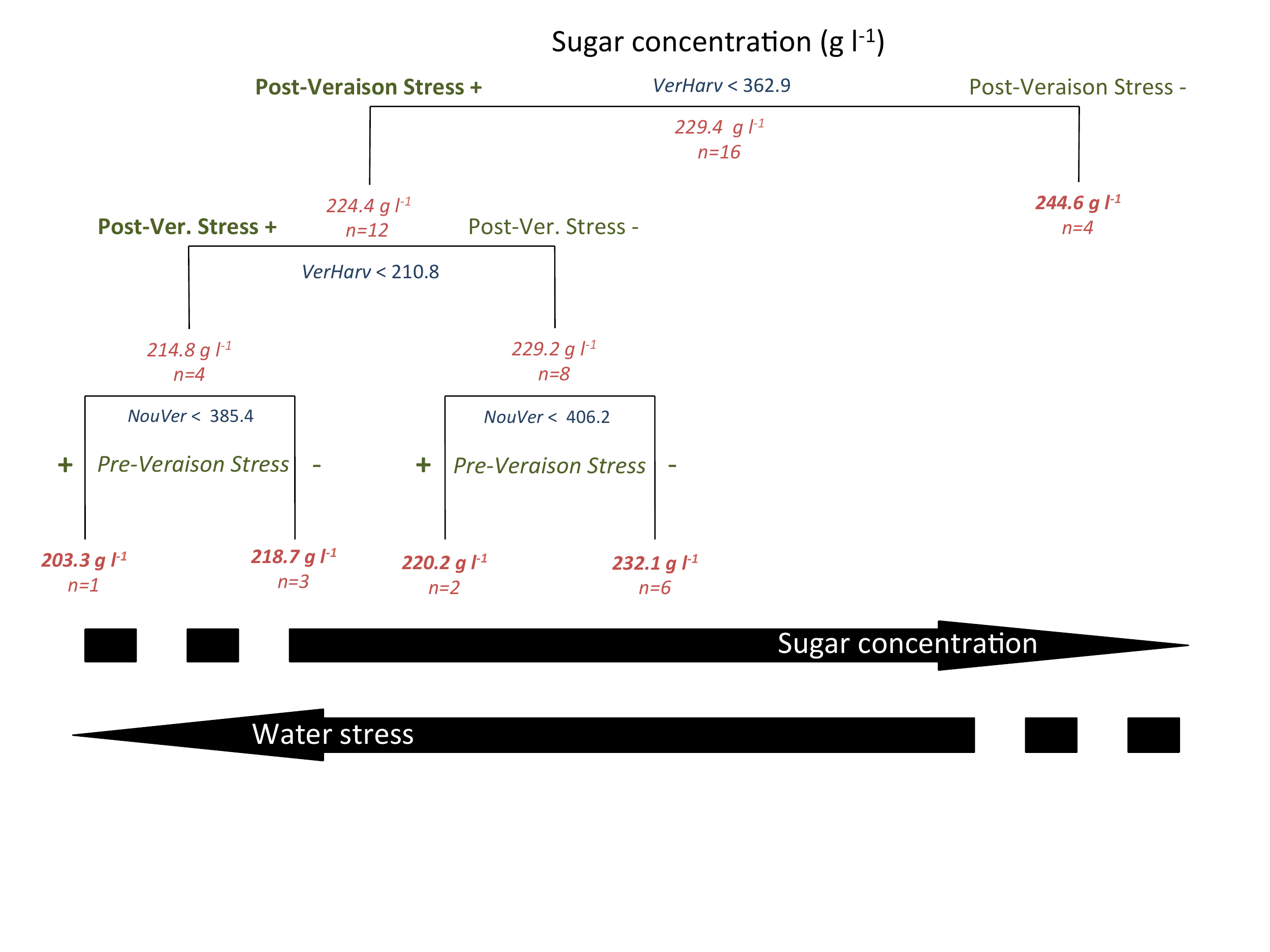}
\caption{Regression tree explaining Sugar Concentration in berries ($\gram \ \reciprocal\litre$) using scalars summarizing the three periods, i.e. pre-veraison (\textit{NouVer}), and post-veraison either until maturity (\textit{VerMat}) or harvest (\textit{VerHarv}). }%Cross validation error was 0.018.}
\label{fig:RTsugar}
\end{figure}

Regarding Sugar Concentration, regression trees show that it is affected by water stress in both pre-veraison \textit{NouVer} and post-veraison \textit{VerHarv} periods (Fig.\ref{fig:RTsugar}). The first discriminant variable on Sugar Concentration is the post-veraison water stress (\textit{VerHarv}, Fig.\ref{fig:RTsugar}). The first split shows that a higher post-veraison water stress leads to a lower Sugar Concentration. 

The left branch resulting from the first split shows that the next discriminant variable is again the post-veraison stress \textit{VerHarv}, so enhancing the effect of the previous split. Lastly, pre-veraison water stress (\textit{NouVer}) can exacerbate the decrease in Sugar Concentration (as shown in the tree bottom). 

%However, while water stress before veraison had no effect on Sugar Concentration in the case of low water stress post-veraison, the effects of water stress pre-veraison can exacerbate the effects of strong water stress pre-veraison, leading to lower Sugar Concentrations in berries (Fig.\ref{fig:RTsugar}).

\subsubsection{Functional data analysis } \label{sec:lfm}

Using a continuous indicator of water deficit enables the use of the whole season water stress curve to explain berry composition. This in turn will promote a more precise monitoring of vine water needs according to the targeted fruit composition. Using the FLRTI method \cite{james2009}, we analyzed the effects of water stress curve over the season on Berry Weight and Sugar Concentration in berry at harvest. 

Regarding Berry Weight, the results showed no significant effect of $\mathit{Ks(t)}$. This was confirmed by applying a specific testing procedure designed for functional linear models, see \cite{hilgert2013}.

\begin{figure}[htbp]
\centering\includegraphics[width=0.99\linewidth]{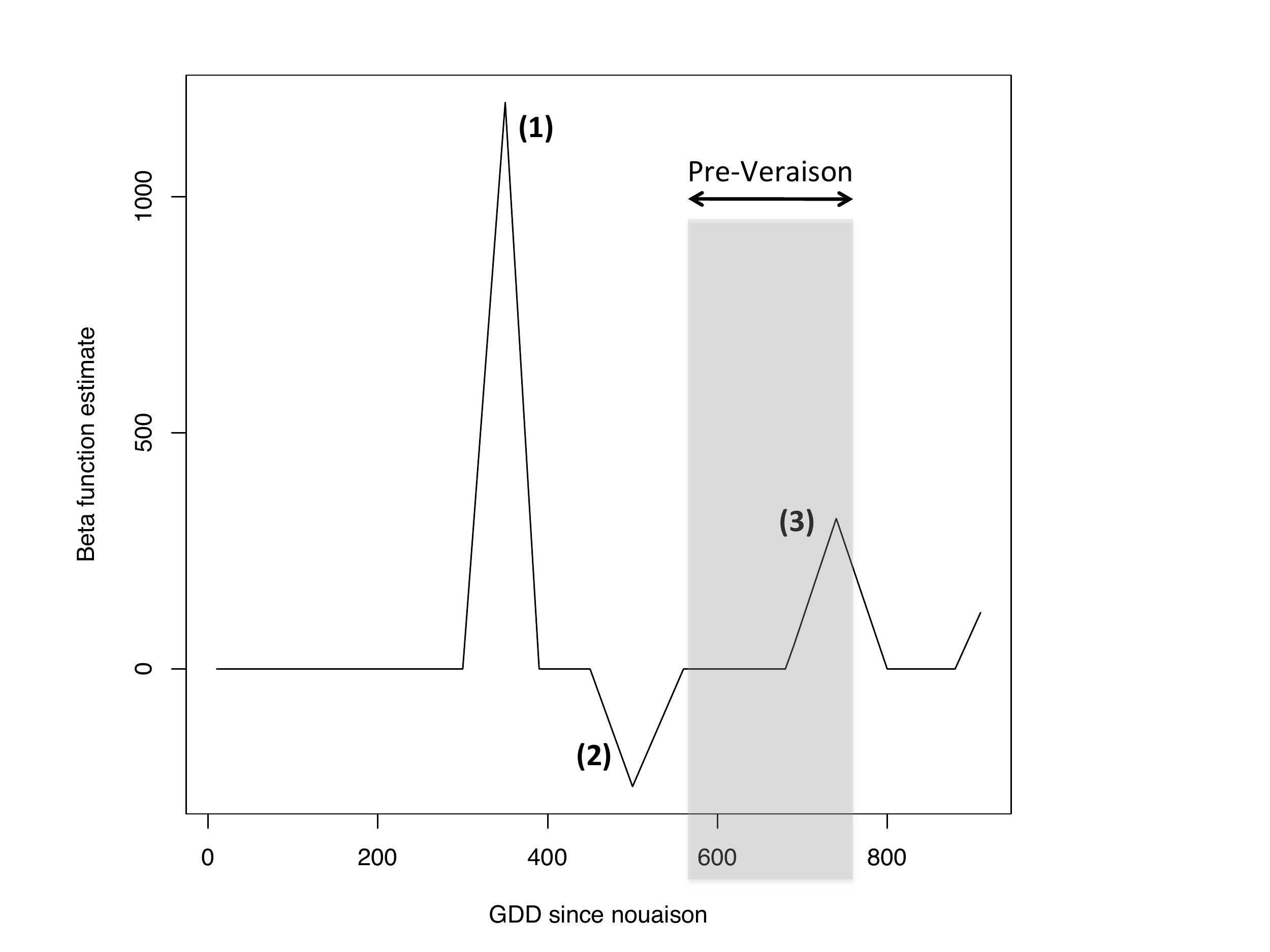}
\caption{Beta Coefficient evolution over time (see Eq. \ref{model-FLR}), for explaining Sugar Concentration at harvest. Abscissa are in GDD. The values of $\sigma=0.05$ and $\omega=0.95$ have been found by cross-validation.}
\label{fig:FuncA}
\end{figure}

Results of the functional data analysis on the effect of $\mathit{Ks(t)}$ on Sugar Concentration are displayed on Fig.\ref{fig:FuncA}. The tuning parameters are indicated in the legend. Parameters were obtained following a ten-fold cross validation. A sensitivity analysis to small $\sigma$ and $\omega$ variations showed a good robustness of the model, with three main peaks always located in the same time periods across the different varietals. These three main peaks are labeled (1), (2) and (3). Each of them corresponds to a significant effect of $\mathit{Ks}$ on Sugar Concentration which can be positive (peaks (1) and (3)) or negative (peak (2)).

Peaks (1) and (3) are positive, which implies a rise in Sugar Concentration. During these periods, the stronger the $\mathit{Ks}$ value, the higher the rise. As $\mathit{Ks}$ varies inversely with water stress, it means that the lower the water stress during these periods, the higher the rise in Sugar Concentration. The effect is twice as strong  for the peak (1) than for the peak (3). 

Regarding the time period, peak (1) appears to be located before pre-veraison whereas peak (3) occurred during pre-veraison. By contrast, peak (2) has a negative effect on Sugar Concentration. During this period located just before pre-veraison, a low water stress decreases the Sugar Concentration. This can be reformulated as follows: the higher the stress during that period, the lower the decrease in Sugar Concentration.

\section{Conclusion}\label{sec:discu}
The work presented in this paper used formalized knowledge and mathematical models to design a software sensor from raw data and relate its temporal output to product quality. The proposed approach has been applied to the case of a vine water stress indicator, and its relation to two grape quality variables: Berry Weight and Sugar Concentration.

Results provide a number of meaningful insights. 

First of all, the software sensor key point, which is the determination of $K^*$, seems reasonably consistent with the literature. From an agronomical point of view, this allows to effectively work at plot scale, and to offer decision support for irrigation, as a function of each plot characteristics.

The use of an ontology allows to separate expert knowledge and numerical models. It makes it much easier to build a generic model, that is both evolutive and adaptable over time as knowledge progresses or climate changes. 

Contrary to a data base, an ontology schema adds semantics to the data structure, allowing automatic reasoning, using logical properties, such as reflexivity or transivity. 

The ontology presented here has a moderate complexity level: only four kinds of primary concepts, and five types of relations. This is still sufficient to express many mathematical conditions and dependencies, going well beyond the scope of the present case study. 
There may however be cases where new  concepts and relations are necessary, and the ontology can easily be enriched when needed.

\vspace{0.2cm}

Second, the two-fold proposal for data analysis appears to be a good means of exploiting such temporal data as provided by the water stress indicator $\mathit{Ks(t)}$. The results show that the water stress has an effect on grape quality. The analysis confirmed known results on the vine physiological response  according to the variety and the irrigation effect. Thus our results are comforting the validity of the  $\mathit{Ks}$ indicator, and therefore the level of confidence and reliability in the software sensor design procedure. Functional data analysis highlighted critical periods for vine and berry development, regarding final quality features.

On one hand, the knowledge-based extraction of meaningful summary features over phenological periods of interest allowed to feed these features as input to decision trees. This confirmed the primordial effect played by the variety on Berry Weight determination.
On the other hand, functional data analysis made it possible to use  the  water stress curve ($\mathit{Ks(t)}$), as a whole, to explain  Sugar Concentration. This allows more precise monitoring of vine water needs according to a targeted product. 

Note that we did not take account of the variety factor in functional data analysis. This would require a covariance analysis model adapted to functional data, which was not possible in this study as the number of data per variety is not sufficient. 

These results show the complementarity of both approaches: the first one performs dimensional reduction by summarizing features which requires expert assumptions, the second one handles the continuous temporal data, without any reduction, but it needs more numerous data to be efficient. 

\vspace{0.2cm}

%le fait de pouvoir utiliser les flux de sève à travers une information continue (courbes) ou ponctuelle (variables agrégées). Le fait de réussir à bien séparer la connaissance experte du reste. Je pense qu'on peut ouvrir dessus en disant qu'on abouti alors à un modèle générique adaptable en fonction des avancées de la connaissance ou du climat. Thibaut voulait insister aussi sur le fait qu'on arrivait à un K* par parcelle ce qui permettra, je pense, d'ouvrir sur le fait qu'on peut vraiment travailler à l’échelle de la parcelle et adapter les préconisations en fonction des caractéristiques de cette parcelle.

Applied perspectives of this work include the study of the relationship between vine water stress and other more complex quality features. In particular new chemical analyses make it possible to follow the aroma development in berries over time, which is assumed to be very sensitive to the grape water status.

Our approach is innovative in more than one aspect. Even if the software sensor had a different design, the same advanced methodology could still be applied to analyze the temporal data.
Beyond the present case study, the proposed methodology has a high genericity level, for the applied fields of Agronomy and Environment. It could be used in many cases when raw data have to be transformed by software sensors to be meaningful, or when temporal data have to be analyzed in depth.

\section*{Acknowledgments}
The research leading to these results has received funding from the Pilotype Program, funded by OSEO innovation and the Languedoc Roussillon regional council. The authors would like to thank all members of the project for their help and advices: Les grands Chais de France, Alliance Minervois, Les Vignerons du Narbonnais, INRA Pech Rouge, INRA SPO, INRA MISTEA, IFV Rh\^one M\'editerran\'ee, SupAgro Montpellier (ITAP), Fruition Sciences, Nyseos. 

Finally, we wish to particularly thank Nicolas Saurin (INRA), Denis Caboulet, Jean-Christophe Payan and Elian Salan\c{c}on (IFV) for providing the vine and wine-related data.

%\appendix
%\makeatletter
%\def\@seccntformat#1{~\csname the#1\endcsname:\quad}
%\makeatother

\newpage
%\section*{References}
%\bibliography{biblio}
%\bibliographystyle{elsarticle-num-names}
%\bibliographystyle{elsarticle-harv}

\end{document}